%% file: main.tex
\documentclass[runningheads]{llncs}

\usepackage[utf8]{inputenc} 
\usepackage[T1]{fontenc}    
\usepackage{hyperref}       
\usepackage{url}            
\usepackage{booktabs}       
\usepackage{amsfonts}       
\usepackage{nicefrac}       
\usepackage{microtype}      
\usepackage{graphicx}
\usepackage{doi}
\usepackage{textcomp}
\usepackage{multirow}
\usepackage[abs]{overpic}
\usepackage[dvipsnames]{xcolor}
\usepackage{graphicx}
\usepackage{float}
\usepackage{tikz}
\usepackage{amsmath}
\usepackage{pdfpages}
\usepackage{listings}
\usepackage{comment}
\usepackage{amsmath,amssymb}
\usepackage{ftnright}
\usepackage{enumitem}
\usepackage{adjustbox}
\usepackage{diagbox}

\usepackage[
backend=bibtex,
style=numeric,
citestyle=numeric
]{biblatex}
\usepackage[width=122mm,left=12mm,paperwidth=146mm,height=193mm,top=12mm,paperheight=217mm]{geometry}

\newcommand{\B}[1]{\textbf{#1}}

\begin{document}
\pagestyle{headings}
\mainmatter
\def\ECCVSubNumber{6083}  

\title{MDMMT-2: Multidomain Multimodal Transformer for Video Retrieval, One More Step Towards Generalization}

\author{Alexander Kunitsyn \and
Maksim Kalashnikov \and \\
Maksim Dzabraev \and
Andrei Ivaniuta}
\authorrunning{A. Kunitsyn et al.}
\titlerunning{MDMMT-2}
%
\institute{Huawei\\
\email{\{kunitsyn.alexnder, maxim.kalashnikov, \\ dzabraev.maksim1, ivanyuta.andrey\}@huawei.com}}

\maketitle

%

\begin{abstract}
In this work we present a new State-of-The-Art on the text-to-video retrieval task on MSR-VTT, LSMDC, MSVD, YouCook2 and TGIF obtained by a single model. Three different data sources are combined: weakly-supervised videos, crowd-labeled text-image pairs and text-video pairs. A careful analysis of available pre-trained networks helps to choose the best prior-knowledge ones. We introduce three-stage training procedure that provides high transfer knowledge efficiency and allows to use noisy datasets during training without prior knowledge degradation. Additionally, double positional encoding is used for better fusion of different modalities and a simple method for non-square inputs processing is suggested.

\end{abstract}

\section{Introduction} \label{sec:introduction}

\input{sect_introduction.tex}

\section{Related work}

\input{sect_related_work.tex}

\section{Methodology}
\input{sect_methodology.tex}

\section{Experiments}
\input{sect_experiments.tex}

\section{Conclusions}
\input{sect_conclusions.tex}

\clearpage
\sloppy
\printbibliography

\end{document}

%% file: sect_introduction.tex
The text-to-video retrieval task is defined as search of most relevant video segments for an arbitrary natural language text query. A search query may contain description of arbitrary actions, objects, sounds or a combinations of them. Note that an arbitrary search query means zero-shot mode of search. A specific search query might not occur in the training database. Despite this, the model should successfully perform the search operation.

The text-to-video retrieval technology can be used for semantic search within a single long video. For example, inside a full-length movie or a stream video. After describing the event, the user can easily find the appropriate video segment. A more general task is the search for a relevant video segment within a large gallery, for example, the entire video hosting like YouTube or Vimeo.

Another application is the search for a specific event in a surveillance cameras dataset or real time video stream. This can be useful to identify illegal actions, accidents or any other important events.

An important requirement for a text-to-video retrieval system is scaling to a large video gallery. A good example of an efficient architecture is the two-stream models. Within this approach the video segment and the text query are encoded independently by the video model and text model respectively. Separate processing allows to compute embeddings for the entire video gallery beforehand. During the inference time, the system calculates the embedding for the search query. Next it calculates the similarity function between query embedding and each embedding from the gallery. Most common choice for similarity function is the cosine similarity.

The data required for the training consists of pairs of video segment and text description. Noise Contrastive Estimation (NCE) is currently the most common framework for this task~\cite{gabeur2020multimodal, miech19howto100m, portilloquintero2021straightforward, CLIP4Clip, clip2video, camoe, gao2021clip2tv}. Within the framework, the model learns to distinguish a positive pair from a set of negative pairs. The most popular losses used in NCE are bi-directional max-margin ranking loss~\cite{Karpathy2014DeepFE} and symmetric cross entropy loss~\cite{NIPS2016_6b180037,oord,infoNCE}.

Since a search query may describe a sound or a visual component, it is important to capture information from both visual stream and audio stream of input video. In this work we fuse information from three modalities: RGB modality (processes each frame independently), motion modality (processes multiple consecutive frames) and audio modality.

%% file: sect_related_work.tex
The text-to-video retrieval task originates in 2016 from work~\cite{rohrbach2016movie}.

Nowadays there is large a number of high-quality crowd-labeled datasets suitable for text-to-video retrieval task~\cite{xu2016msr-vtt,caba2015activitynet,tgif-cvpr2016,rohrbach2016movie,chen-dolan-2011-collecting,wang2020vatex,lei2019tvqa,2020trecvidawad,ZhXuCoAAAI18,goyal2017something} and numerous works using these datasets~\cite{gao2021clip2tv,camoe,clip2video,CLIP4Clip,laff,mdmmt,gabeur2020multimodal,yang2021taco}. In~\cite{miech2020endtoend} the authors leverage large amount of weakly-supervised data (HT100M dataset) from YouTube to train a model. In~\cite{gabeur2020multimodal,mdmmt} both weakly-supervised data for pre-training and crowd-labeled datasets for fine-tuning are used.

The task requires a large amount of data and looking for alternative data sources is quite reasonable.
Since the visual stream of a video is a sequence of frames (images), any individual image can be considered as a one-frame video. In~\cite{miech2020learning} the authors successfully use both image-text and video-text datasets.

Impressive results are achieved in text-to-image retrieval by CLIP model, which is trained with a large amount of web-crawled data~\cite{radford2021learning}.

To create a text-to-video retrieval model for general application (without specialization for a particular domain), a large amount of data is required. The authors of CLIP use hundreds of millions of data units to train text-to-image retrieval model for general application. Most probably text-to-video retrieval task requires no less and rather more data.

Unfortunately, combining all crowd-labeled text-video and text-image datasets do not allow to approach to the high quality general application model. In~\cite{miech2020endtoend} the authors attempt to use large amount of weakly-supervised data but the result is still far from high quality model. 

Transfer learning-based methods are getting more and more popular to be applied for this task. One of the first successful applications of transfer learning for text-to-video retrieval task can be attributed to~\cite{miech2018learning}, where several pre-trained networks are used to extract features from video. In~\cite{gabeur2020multimodal} the authors additionally adopted BERT model~\cite{bert} as initialization for text encoder. Later works~\cite{gao2021clip2tv,camoe,clip2video,CLIP4Clip,mdmmt} use CLIP model as initialization for both text and vision encoders.

Pre-trained models, suitable for the text-to-video retrieval task can be divided into two classes. The first class is trained using crowd-labeled datasets such as Imagenet~\cite{imagenet_cvpr09} or Kinetics~\cite{kay2017kinetics} datasets. Usually such models produce task-specific embeddings, which does not allow to achieve high quality in the text-to-video retrieval task. The second class is trained with a large amount of weakly-supervised data collected from the Internet. The most popular are CLIP, BERT and irCSN152, which are trained with the IG65M dataset (irCSN152-IG65M)~\cite{ghadiyaram2019largescale}.

The analysis of pre-trained models in~\cite{mdmmt} and our experience show that models trained with a large amount of web-crawled data are able to produce embeddings for general application and allow to reach better quality in the text-to-video retrieval task.

Using CLIP as an initialization or a feature extractor significantly improves the results in the text-to-video retrieval task~\cite{gao2021clip2tv,camoe,clip2video,CLIP4Clip,mdmmt}. The CLIP model family has several different architectures. All of them have independent text encoder and visual encoder. 

In this work we manage to use text-video, text-image and text-video weakly-supervised (HT100M) datasets together in the same training. In addition, we use the best pre-trained models. This allows us to achieve State-of-The-Art results with a single model on a number of benchmarks.

%% file: sect_methodology.tex
Our model follows the idea of MDMMT~\cite{gabeur2020multimodal, mdmmt}. However, we suggest an advanced multistage training approach, as well as perform analysis of existing prior knowledge and choose optimal backbones. 


\subsection{Architecture} \label{ssec:architecture}

The architecture consists of four parts: pre-trained experts, aggregator, text encoder and text embedding projection.

Pre-trained expert is a frozen pre-trained network that produces sequence of features for input video. In this work we use three experts, each for different modality. The first one is for image (RGB modality), processes video frames independently. The second one is for motion. It deals with several continuous frames together. The third one is for audio. See pseudocode example in Lst.~\ref{lst:expert_example}.

The aggregator accepts embeddings made by experts and produce single embedding for the video. See pseudocode example in Lst.~\ref{lst:aggregator_example}.

The text encoder accepts arbitrary English natural language text and produces embedding.
\begin{lstlisting}[label=lst:expert_example, language=Python, caption={Example of pre-trained expert usage},captionpos=t, basicstyle=\ttfamily\small]
def encode_rgb(V):
  # V: input video
  embs = []
  frames_lst = read_1_frame_per_second(V)
  for frame in frames_lst:
    emb = image_network(frame)
    embs.append(emb)
  rgb_embs = concatenate(embs, dim=0)
  return rgb_embs
\end{lstlisting}

\begin{lstlisting}[label=lst:aggregator_example, language=Python, caption={Example of aggregator},captionpos=t, basicstyle=\ttfamily\small]
def aggregator(rgb_embs, motion_embs, audio_embs):
  rgb_embs = FC_768_to_512(rgb_embs)
  rgb_cls = rgb_embs.max(dim=0) + rgb_bias # (1, 512)
  rgb_input = rgb_embs + positional + rgb_bias
  # do the same for other modalities
  x = concatenate([
    rgb_cls, motion_cls, audio_cls,
    rgb_input, motion_input, audio_input], dim=0)
  x = transformer_encoder(x)
  x = normalize(x)
  video_emb = x[:3].reshape(-1) # (512*3, )
  return video_emb
\end{lstlisting}

The text embedding projection part maps text embedding to distinct space for each modality. See example in Lst~\ref{lst:projection_example}. GEU* means Gated Embedding Unit~\cite{liu2020use}.

\begin{lstlisting}[label=lst:projection_example, language=Python, caption={Text embedding projection},captionpos=t, basicstyle=\ttfamily\small]
def text_embedding_projection(temb):
  a1, a2, a3 = softmax(FC_512_to_3(temb))
  temb_rgb = a1 * GEU1(temb)
  temb_motion = a2 * GEU2(temb)
  temb_audio = a3 * GEU3(temb)
  return concatenate([
    temb_rgb, temb_motion, temb_audio
  ]) # (512*3, )
\end{lstlisting}

Note that this architecture is flexible. It is possible to remove or add additional modalities. Also it is possible to replace a given pre-trained text encoder with another one. For example, it is possible to use CLIP ViT-B/32 as RGB expert and text part of CLIP ViT-B/16 as text encoder.

\subsection{Double positional encoding}\label{ssect:double-pos-end}

Each expert takes different type and shape of data as input. For example, CLIP takes a single image frame to produce an embedding. irCSN152-IG65M produces a single embedding from a sequence of 32 consecutive frames. Slow-Fast (SF)~\cite{kazakos2021slowfast} takes a melspectrogram of a 5 seconds long audio frame to produce an embedding. 

Positional encoding is used in the transformer encoder architecture to provide information about the order of tokens in input sequence. In our case, positional (temporal) encoding has to provide information not only about the order of tokens, but also about the time length of each individual token. 



We introduce double positional encoding. For each embedding we add two biases: the first stands for the timestamp of beginning of video segment and the second represents timestamp of end of video segment, see pseudocode in Lst.~\ref{lst:double-pos-emb}.

\begin{lstlisting}[label=lst:double-pos-emb, language=Python, caption={Pseudocode for double positional encoding},captionpos=t, basicstyle=\ttfamily\small]
# nsec: video duration in seconds
positions_beg = nn.Parameter(32, 512)
positions_end = nn.Parameter(32, 512)
audio_embs = audio_embs + 
             positions_beg[0::5][:nsec//5] +
             positions_end[5::5][:nsec//5]
rgb_embs = rgb_embs +
           positions_beg[:nsec] +
           positions_end[1:][:nsec]
\end{lstlisting}

This way we make sure that different time lengths per expert embedding are processed correctly. The results in Tab.~\ref{tab:temporal_embedding}  support this novelty.
\begin{table}[t]
	\centering
	\caption{Comparison of standard positional encoding with proposed double positional encoding. Dataset: MSR-VTT full clean split (see Sec.~\ref{ssec:datasets}); Text backbone: CLIP ViT-B/32; Experts: CLIP ViT-L/14, irCSN152-IG65M, SF}
	\label{tab:temporal_embedding}
	
	\begin{tabular}{|c|l @{\hspace{1\tabcolsep}} l @{\hspace{1\tabcolsep}} l @{\hspace{1\tabcolsep}}l|}
		\toprule
		Temporal  & \multicolumn{4}{c|}{Text $ \rightarrow$ Video} \\
		Embedding	  & R@1$\uparrow$ & R@5$\uparrow$ & R@10$\uparrow$ & MdR$\downarrow$ \\
		\midrule
		Single   & 22.1$_{\pm 0.1}$ & 48.2$_{\pm 0.0}$   & 60.0$_{\pm 0.1}$ & 6.0$_{\pm 0.0}$\\
		Double   & 22.2$_{\pm 0.1}$ & 48.5$_{\pm 0.2}$   & 60.3$_{\pm 0.2}$ & 6.0$_{\pm 0.0}$\\
		\bottomrule
	\end{tabular}	
\end{table}

\subsection{Datasets} \label{ssec:datasets}
A list of datasets used in this work is provided in Tab.~\ref{tab:dataset_size}. Only training splits of listed datasets are used in training dataset. Note that we use both text-video and text-image datasets. In Sec.~\ref{ssec:adding_images} we show results for video only datasets and image plus video datasets. Since each dataset has different amount of videos and captions, it is important to combine datasets properly~\cite{mdmmt}. 

In the following experiments MSR-VTT full clean split is used. This split is introduced in~\cite{mdmmt}. The test part of full clean split is the same as test part of full split. The training part of full clean split mostly similar to full split, but some videos are removed. All removed videos have corresponding duplicate in test part.

\begin{table}[t]
	\centering
		\caption{The "Num videos" column represents the number of video clips (images) in the dataset,
			 the "Num pairs" column represents the total number of video-caption (image-caption) pairs,
			 the "Num unique captions" column represents the number of unique captions in the dataset}
	    \label{tab:dataset_size}
	    
	\begin{tabular}{ |c|ccc| }
		\toprule
		\multirow{3}{*}{Dataset}  & Num   & Num   & Num       \\
								  & videos & pairs & unique    \\
								  & (images)      &       & captions  \\
		\midrule
		MSR-VTT~\cite{xu2016msr-vtt}	& 10k		    & 200k			& 167k  \\
		ActivityNet~\cite{caba2015activitynet}		& 14k		    & 70k			& 69k   \\
		LSMDC~\cite{rohrbach2016movie}			& 101k		    & 101k			& 101k  \\
		TwitterVines~\cite{2020trecvidawad}    & 6.5k		    & 23k			& 23k   \\
		YouCook2~\cite{ZhXuCoAAAI18}		& 1.5k		    & 12k			& 12k   \\
		MSVD~\cite{chen-dolan-2011-collecting}			& 2k		    & 80k			& 64k   \\
		TGIF~\cite{tgif-cvpr2016}			& 102k          & 125k          & 125k  \\
		SomethingV2~\cite{goyal2017something}		& 193k		    & 193k			& 124k  \\
	    VATEX~\cite{wang2020vatex}           & 28k		    & 278k			& 278k  \\
	    TVQA~\cite{lei2019tvqa}            & 20k		    & 179k			& 178k  \\
	    \textit{\bf Sum above}& {\it\bf 477k} & {\it\bf 1261k}  &\\
	    \midrule
	    Flicker30k~\cite{young-etal-2014-image}         & 32k		    & 159k			& 158k  \\
	    COCO~\cite{chen2015microsoft}            & 123k		    & 617k			& 592k  \\
	    Conceptual Captions~\cite{Sharma2018ConceptualCA}    & 3M		    & 3M			& 2M  \\
		\bottomrule
	\end{tabular}
\end{table}

\subsection{Loss} \label{ssec:loss}
The MDMMT-2 is trained with the bi-directional max-margin ranking loss~\cite{Karpathy2014DeepFE}:
\begin{equation}
	\frac{1}{B}\sum_{i=1}^{B} \sum_{j \neq i} \Big[ \max(0, s_{ij} - s_{ii} + m) + \max(0, s_{ji} - s_{ii} + m)\Big]
\end{equation}
where $B, s_{ij}, m$ denote the batch size, the similarity score between the $i$-th query and the $j$-th video of the given batch, and some predefined margin correspondingly. We set $m=0.05$ and $B=256$ in all our experiments.

%% file: sect_experiments.tex
In sections~\ref{ssec:clip} - \ref{ssec:non_square} all experiments are made on MSR-VTT full clean split (see Sec.~\ref{ssec:datasets}) for 50 epochs and 60k examples per epoch. The initial learning rate is 5e-5. After each epoch we multiply learning rate by $\gamma=0.95$. In these experiments we freeze text backbone and train only aggregator model and text embedding projection part. 

For training on MSR-VTT, we use aggregator with 4 layers and 4 heads. On larger dataset (see Sec.~\ref{ssec:adding_images} - \ref{ssec:final_result}) aggregator has 9 layers and 8 heads.

Results are reported as \textit{mean}$_{\pm \textit{std}}$ or just \textit{mean} over 3 experiments.

\subsection{CLIP} \label{ssec:clip}
In~\cite{mdmmt} it is shown that CLIP works as a strong visual feature extractor and outperforms other available models by large margin. We found out that CLIP text backbone also works better than other available text models, such as BERT~\cite{bert}, which was originally used in~\cite{gabeur2020multimodal}, or GPT~\cite{gpt}.

Currently there are several publicly available CLIP models. In this section we compare their performance to make sure that we use the best possible combination. Results are presented in Tab.~\ref{tab:clip}.

Our observations:
\begin{itemize}
    \item Suppose we have pre-trained CLIP: text backbone and corresponding visual backbone. We observe that if we replace original visual backbone with
a bigger/deeper one, we obtain better video retrieval system.
    \item If we use the same visual backbone with different text backbones, a text backbone of a bigger/deeper model not necessarily shows better results. \\
    In fact, if we take a look at (Tab.~\ref{tab:clip}) RN50(xN) models, the best result is achieved by a combination of the deepest visual backbone (RN50x64) and the text backbone from the most shallow model (RN50).
    \item CLIP ViT-L/14 shows the best performance both as visual and text backbone.
    
\end{itemize}

\begin{table}
	\centering
	\caption{Comparison of CLIP visual and text backbones combinations. Experts: CLIP; Metric: R@5}
	\label{tab:clip}
	\begin{tabular}{|c|ccccccc|}
         \toprule 
		 \diagbox{Visual}{\rotatebox{90}{Text}}&
		 \rotatebox{90}{RN50} & \rotatebox{90}{RN50x4} & \rotatebox{90}{RN50x16}  & \rotatebox{90}{RN50x64} & \rotatebox{90}{ViT-B/32} &  \rotatebox{90}{ViT-B/16} & \rotatebox{90}{ViT-L/14} \\
		\midrule
		 RN50			& 40.1 & 38.7 & 39.3 & 39.3 & 40.1 & 39.8 & 39.8  \\
		 RN50x4			& 42.8 & 41.9 & 42.5 & 42.5 & 43.2 & 43.1 & 43.2   \\
		 RN50x16		& 43.9 & 43.5 & 43.6 & 43.0 & 44.4 & 44.5 & 44.4   \\
	     RN50x64		& 44.6 & 43.9 & 44.1 & 44.2 & 44.8 & 45.2 & 45.4   \\
		 ViT-B/32		& 42.0 & 41.2 & 40.9 & 40.9 & 42.5 & 42.4 & 42.2   \\
		 ViT-B/16		& 44.4 & 43.8 & 43.4 & 43.3 & 44.8 & 45.4 & 44.9   \\
		 ViT-L/14		& 46.2 & 45.7 & 45.3 & 45.3 & 46.5 & 46.8 & \B{47.2}   \\
		\bottomrule
	\end{tabular}
\end{table}

\begin{table}
	\centering
	\caption{ Experiments on different audio experts. Text backbone: CLIP ViT-B/32; Experts: CLIP ViT-B/32, irCSN152-IG65M, audio}
	\label{tab:audio_experts}	

	\begin{tabular}{|c|llll|}
		\toprule
			Audio  & \multicolumn{4}{c|}{Text $ \rightarrow$ Video} \\
			expert	  & R@1$\uparrow$ & R@5$\uparrow$ & R@10$\uparrow$ & MdR$\downarrow$ \\
		\midrule
		vggish~\cite{hershey2017cnn}   & 19.3$_{\pm 0.2}$  & 44.3$_{\pm 0.0}$    & 56.3$_{\pm 0.2}$ & 7.0$_{\pm 0.0}$ \\
		Slow-Fast~\cite{kazakos2021slowfast} & 19.6$_{\pm 0.3}$  & 44.9$_{\pm 0.3}$    & 57.0$_{\pm 0.2}$ & 7.0$_{\pm 0.0}$ \\
		\bottomrule
	\end{tabular}		
\end{table}

\subsection{Experts combination} \label{ssec:experts_combination}
\begin{table}
	\centering
	\caption{Experts combinations. Text backbone: CLIP ViT-B/32}
	\label{tab:experts_combination}
	
	\begin{tabular}{|ccc|lll|}
		\toprule
		\multicolumn{3}{|c|}{Experts} & \multicolumn{3}{c|}{Text $ \rightarrow$ Video} \\ 
		CLIP & irCSN152-IG65M & SF & R@1$\uparrow$ & R@5$\uparrow$ & MdR$\downarrow$ \\
		\midrule
		& \checkmark &                       
		& 10.2$_{\pm 0.0}$  & 29.3$_{\pm 0.1}$ & 17.3$_{\pm 0.5}$\\
	    & \checkmark & \checkmark                    
		& 11.2$_{\pm 0.1}$  & 31.5$_{\pm 0.2}$ & 15.0$_{\pm 0.0}$\\
		\checkmark  &  &                        
		& 21.3$_{\pm 0.1}$  & 46.5$_{\pm 0.2}$ & 7.0$_{\pm 0.0}$\\
		\checkmark    & \checkmark &            
		& 21.5$_{\pm 0.1}$  & 46.7$_{\pm 0.1}$ & 7.0$_{\pm 0.0}$\\
		\checkmark     &  & \checkmark          
		& 22.0$_{\pm 0.1}$  & 47.8$_{\pm 0.1}$ & 6.0$_{\pm 0.0}$\\
		\checkmark    & \checkmark & \checkmark 
		& \B{22.2}$_{\pm 0.1}$  & \B{48.5}$_{\pm 0.2}$ & \B{6.0}$_{\pm 0.0}$\\
		\bottomrule
	\end{tabular}	
\end{table}
Using combination of different experts allows to achieve better performance. In Tab.~\ref{tab:experts_combination} various combinations of experts are presented. Using three modalities gives the best result.
\subsection{Dealing with non-square videos} \label{ssec:non_square}
\begin{table}
	\centering
	\caption{
 Comparison of different techniques for extracting features from non-square videos. Text backbone: CLIP ViT-B/32; Experts: CLIP ViT-L/14; Metric: R@5}
	\label{tab:crop}
	
	\begin{tabular}{|c|*{4}c|}
		\toprule
		 \diagbox{Train}{Test}&Squeeze &  Center crop & Padding  & Mean \\
		\midrule
		Squeeze     & 46.3 & 46.0 & 46.0 & \textbf{47.1} \\
		Center Crop & 46.0 & 46.5 & 46.0 & \textbf{47.3} \\
		Padding     & 46.0 & 46.2 & 46.7 & \textbf{47.0} \\  
	    Mean        & 45.9 & 46.4 & 45.9 & \textbf{47.4} \\                
		\bottomrule
	\end{tabular}
\end{table}
Both irCSN152-IG65M and CLIP take videos (images) of square shape as input. Therefore it is not possible to use information from the whole video directly. It may happen that some object or action is taking place in the corner (out of the center crop) of the video. So if we use center crop to compute embeddings, the information from the corners will be lost. There are several possible solutions to this problem:
\begin{itemize}
    \item Squeeze a video to a square without saving the aspect ratio (\textit{squeeze})
    \item Pad a video to a square with blackbars (\textit{padding})
    \item Take several crops from the video, average the embeddings of these crops, and use this average as embedding (\textit{mean}) 
\end{itemize}

For the \textit{mean} technique we take three crops: left or bottom, center, right or top (depending on video orientation) and then average embeddings of these crops.

Experiments in Tab.~\ref{tab:crop} show that
\textit{squeeze} works worse than center crop,
\textit{padding} works slightly better than center crop, and \textit{mean} works the best.

We want to emphasize that using \textit{mean} during test improves video-retrieval performance even if other methods were used during train.

\subsection{Adding images} \label{ssec:adding_images}
\begin{table}
      \centering
      \caption{Datasets used in train procedure. The "Weight" column describes how often we sample examples from the dataset. The probability of obtaining an example from the
      dataset with the weight $w$ equals to $w$ divided by a sum of all weights}
      \label{tab:dataset-wgh}
      \begin{tabular}{|c|c|c|}
	    \toprule
	    Dataset	    & Weight & Type \\
	    \midrule
	    MSR-VTT	        & 140 & \multirow{10}{*}{\shortstack{Text-video\\ datasets \\ (10V)}}\\
	    ActivityNet	    & 100 &\\
	    LSMDC	        & 70  &\\
	    Twitter Vines   & 60  &\\
	    YouCook2	    & 20  &\\
	    MSVD	        & 20  &\\
	    TGIF	        & 102 &\\
	    SomethingV2     & 169 &\\
	    VATEX           & 260 &\\
	    TVQA            & 150 &\\
	    \midrule
	    COCO            & 280 &\multirow{3}{*}{\shortstack{Text-image\\ datasets \\ (3I)}}\\
	    Flicker30k      & 200 &\\
	    Conceptual Captions    & 160 &\\
	    \bottomrule
      \end{tabular}
\end{table}

\begin{table}
  \centering
  \caption{ Test results on MSR-VTT full clean split. Text backbone: CLIP ViT-B/32; Experts: CLIP ViT-L/14, irCSN152-IG65M, SF}
  \label{tab:10v}
  
  \begin{tabular}{|c|*{4}c|}
		  \toprule
	       \multirow{2}{*}{Dataset}  & \multicolumn{4}{c|}{Text $ \rightarrow$ Video} \\ 
		 & R@1$\uparrow$ & R@5$\uparrow$ & R@10$\uparrow$ & MdR$\downarrow$ \\
		  \midrule
    10V     &  30.2    & 56.6 & 67.1  &  4.0    \\
    10V+3I  &  30.9    & 57.4 & 67.8  &  4.0     \\
		  \bottomrule
	    \end{tabular}
\end{table}
In \cite{mdmmt} it is shown that the proper combination of datasets allows to train a single model that can capture the knowledge from all used datasets and in most cases the 
model trained on the combination of datasets is better than the model trained on a single dataset. 

In Tab.~\ref{tab:10v} we show that proper combination of text-video and text-image datasets allows to improve video-retrieval performance. Hyperparameters are specified in Sec.~\ref{ssec:stages}, stage S$_1$.

Weights for combining all datasets are specified in Tab.~\ref{tab:dataset-wgh}. First 10 rows are video datasets (denoted as 10V) and last 3 are image datasets (denoted as 3I). 

\subsection{Pre-training and fine-tuning}\label{ssec:stages}
Note that in our work aggregator is initialised form scratch, while text backbone is pre-trained. If we simultaneously train randomly initialised aggregator and pre-trained text backbone, then at the time when aggregator will be trained, the text backbone might degrade. That is why for final result we introduce training procedure that consists of three stages (denoted as S$_0$, S$_1$, S$_2$).

During stage S$_0$ we use noisy HT100M dataset. Text backbone is frozen, only aggregator and text embedding projection part are trained.

During stage S$_1$ we use crowd-labeled datasets 10V+3I . Same as in S$_0$, text backbone is frozen, only aggregator and text embedding projection part are trained.

During stage S$_2$, same as in S$_1$, we use  crowd-labeled datasets 10V+3I. Now, however, we unfreeze text backbone and train all three main components: aggregator, text backbone and text embedding projection.

Hyperparameters for these stages are listed in Tab.~\ref{tab:stages_params}.
Results for different combinations of stages are listed in Tab.~\ref{tab:stages_results}.

\begin{table}
	\centering
	\caption{Hyperparameters for different stages}
	\label{tab:stages_params}
	
	\begin{tabular}{|c|*{4}c|c|}
		\toprule
		Train & Examples & Num. & Learning& \multirow{2}{*}{$\gamma$} & \multirow{2}{*}{Datasets}  \\
		stage&per epoch & epochs & rate && \\
		\midrule
		S$_0$ & 60k & 200 & 5e-5 & 0.98 & HT100M \\
		S$_1$ & 380k & 45 & 5e-5 & 0.95 & 10V+3I \\
		S$_2$ & 200k & 20 & 2e-5 & 0.8  & 10V+3I  \\
		\bottomrule
	\end{tabular}	
\end{table}

\begin{table}
	\centering
	\caption{ Test results for train stages on MSR-VTT full clean split.
  Text backbone: CLIP ViT-B/32; Experts: CLIP ViT-L/14, irCSN152-IG65M, SF}
	\label{tab:stages_results}
	
	\begin{tabular}{|*{3}c|*3{c}|}
		\toprule
		\multicolumn{3}{|c|}{Train stages} & \multicolumn{3}{c|}{Text $\rightarrow$ Video} \\ 
		S$_0$ & S$_1$ & S$_2$ & R@1$\uparrow$ & R@5$\uparrow$ & MdR$\downarrow$ \\
		\midrule
		\checkmark &  &                       
		& 7.7  & 19.0 & 60.0 \\
	    & \checkmark &                   
		& 29.0 & 55.3 & 4.0  \\
		& \checkmark &  \checkmark        
		& 30.5 & 56.9 & 4.0  \\
		\checkmark  & \checkmark   &                        
		& 31.2  & 57.8 & 4.0 \\
		\checkmark    & \checkmark &  \checkmark        
		& 32.5 & 59.4 & 3.0  \\
		\bottomrule
	\end{tabular}	
\end{table}

\subsection{Final result} \label{ssec:final_result}
In this section we compare our solution with the prior art. Our best solution uses three modalities: CLIP ViT-L/14 (RGB modality), irCSN152-IG65M (motion modality), Slow-Fast trained on VGG-Sound (audio modality). Text backbone is used from CLIP ViT-L/14. To fuse modalities
we use aggregator with 9 layers and 8 heads. Training procedure is described in Sec.~\ref{ssec:stages}. Results are shown in Tab.~\ref{tab:models-msrvtt-1kA} - Tab.~\ref{tab:models-tgif}.

Center crop is used for visual features extraction during training and testing for all datasets except MSR-VTT (see Tab.~\ref{tab:models-msrvtt}), where we report two results on testing set: center crop and \textit{mean} method (see Sec.~\ref{ssec:non_square}).

Results on MSR-VTT, LSMDC, MSVD, YouCook2, TGIF are obtained using single model. Our model outperforms SOTA by 1.6, 0.6, 3.9, 4.3, 1.1 \% correspondingly on R@5. On MSR-VTT-1kA (see Tab.~\ref{tab:models-msrvtt-1kA}) we report two results with different training splits: full(7k) and 1k-A(9k). First result approaches SOTA and second result outperforms SOTA by 0.8 \% on R@5.

\begin{table}
  \centering
  \caption{Test results on MSR-VTT-1k-A dataset. Results that were obtained using original testing protocol (without dual softmax~\cite{camoe, gao2021clip2tv} on inference) are shown. Results are collected from articles and \url{https://paperswithcode.com/sota/video-retrieval-on-msr-vtt-1ka}}
  \label{tab:models-msrvtt-1kA}

\begin{tabular}{|l|*{5}l|}
    \toprule
    \multirow{2}{*}{Model} & \multicolumn{5}{c|}{MSR-VTT-1k-A text $\rightarrow$ video} \\
			   & R@1$\uparrow$ & R@5$\uparrow$ & R@10$\uparrow$ & MnR$\downarrow$ & MdR$\downarrow$ \\
    \midrule
JSFusion~\cite{yu2018joint}
& 10.2 & 31.2 & 43.2 & --- & 13.0 \\
E2E~\cite{miech2020endtoend}
& 9.9 & 24.0 & 32.4 & --- & 29.5 \\
HT~\cite{miech19howto100m}
& 14.9& 40.2 & 52.8 & --- & 9.0 \\
CE~\cite{liu2020use}
& 20.9 & 48.8 & 62.4 & 28.2 & 6.0 \\
CLIP~\cite{radford2learning}
& 22.5 & 44.3 & 53.7 & 61.7 & 8.0 \\
MMT ~\cite{gabeur2020multimodal}
& 26.6   & 57.1   & 69.6   & 24.0   & 4.0 \\
AVLnet\cite{rouditchenko2020avlnet}
& 27.1 & 55.6 & 66.6 & --- & 4.0 \\
SSB~\cite{patrick2021supportset}
& 30.1 & 58.5 & 69.3 & --- & 3.0 \\
CLIP agg~\cite{portilloquintero2021straightforward}
& 31.2 & 53.7 & 64.2 & --- & 4.0 \\
MDMMT~\cite{mdmmt}
& 38.9 & 69.0 & 79.7 & 16.5 & 2.0 \\
CLIP4Clip~\cite{CLIP4Clip}
& 44.5 & 71.4 & 81.6 & 15.3 & 2.0 \\
CLIP2Video~\cite{clip2video}
& 45.6 & 72.6 & 81.7 & 14.6 & 2.0 \\
LAFF~\cite{laff}
& 45.8 & 71.5 & 82.0 & --- & --- \\
CAMoE~\cite{camoe}
& 44.6 & 72.6 & 81.8 & \B{13.3} & 2.0 \\
MDMMT-2 full (Ours)
& 46.5$_{\pm 0.8}$ & 74.3$_{\pm 0.6}$ & \B{83.3}$_{\pm 0.2}$ & 14.1$_{\pm 0.1}$ & 2.0$_{\pm 0.0}$ \\
QB-Norm+CLIP2Video~\cite{qbnorm}
& 47.2 & 73.0 & 83.0 & --- & 2.0 \\
CLIP2TV~\cite{gao2021clip2tv}
& 48.3 & 74.6 & 82.8 & 14.9 & 2.0 \\
MDMMT-2 1k-A (Ours)
& \B{48.5}$_{\pm 0.3}$ & \B{75.4}$_{\pm 0.3}$ & \B{83.9}$_{\pm 0.5}$ & 13.8$_{\pm 0.3}$ & \B{2.0}$_{\pm 0.0}$ \\
    \bottomrule
  \end{tabular}
\end{table}

\begin{table}
  \centering
  \caption{Test results on MSR-VTT dataset. Results are collected from articles and \url{https://paperswithcode.com/sota/video-retrieval-on-msr-vtt}}
  \label{tab:models-msrvtt}

  \begin{tabular}{|l|c|*{5}l|}
    \toprule
    \multirow{2}{*}{Model} & \multirow{2}{*}{Split} & \multicolumn{5}{c|}{MSR-VTT text $\rightarrow$ video} \\
			   & & R@1$\uparrow$ & R@5$\uparrow$ & R@10$\uparrow$ & MnR$\downarrow$ & MdR$\downarrow$ \\
    \midrule
VSE~\cite{mithun2018learning} & \multirow{17}{*}{full} & 5.0 & 16.4 & 24.6 & --- & 47.0 \\
VSE++~\cite{mithun2018learning}
&& 5.7 & 17.1 & 24.8 & --- & 65.0 \\
Multi Cues~\cite{mithun2018learning}
&& 7.0 & 20.9 & 29.7 & --- & 38.0 \\
W2VV~\cite{Dong_2018}
&& 6.1 & 18.7 & 27.5 & --- & 45.0 \\
Dual Enc.~\cite{dong2019dual}
&& 7.7 & 22.0 & 31.8 & --- & 32.0 \\
CE~\cite{liu2020use}
&& 10.0 & 29.0& 41.2 & 86.8 & 16.0 \\
MMT ~\cite{gabeur2020multimodal}
&& 10.7 &  31.1 & 43.4 &  88.2 & 15.0 \\
CLIP~\cite{radford2learning}
&& 15.1 & 31.8 & 40.4 &  184.2  & 21.0 \\
CLIP agg~\cite{portilloquintero2021straightforward}
&& 21.5 & 41.1 & 50.4 &  --- & 4.0 \\
MDMMT~\cite{mdmmt}
&& 23.1 & 49.8 & 61.8 & 52.8 & 6.0 \\
TACo~\cite{yang2021taco}
&& 24.8 & 52.1 & 64.0 & --- & 5.0 \\
LAFF~\cite{laff}
&& 29.1 & 54.9 & 65.8 & --- & --- \\
CLIP2Video~\cite{clip2video}
&& 29.8 & 55.5 & 66.2 & 45.4 & 4.0 \\
CAMoE~\cite{camoe}
&& 32.9 & 58.3 & 68.4 & 42.6 & 3.0 \\
CLIP2TV~\cite{gao2021clip2tv}
&& 33.1 & 58.9 & 68.9 & 44.7 & 3.0 \\
MDMMT-2 (Ours)
&& \B{33.4}$_{\pm 0.1}$ & \B{60.1}$_{\pm 0.1}$ & \B{70.5}$_{\pm 0.1}$ & \B{39.2}$_{\pm 0.2}$ & \B{3.0}$_{\pm 0.0}$ \\
MDMMT-2 test \textit{mean} (Ours) 
&& \B{33.7}$_{\pm 0.1}$ & \B{60.5}$_{\pm 0.0}$ & \B{70.8}$_{\pm 0.1}$ & \B{37.8}$_{\pm 0.3}$ & \B{3.0}$_{\pm 0.0}$ \\
    \midrule
MMT~\cite{gabeur2020multimodal}
&\multirow{3}{*}{\shortstack{full\\clean}}& 10.4 & 30.2 & 42.3 & 89.4 & 16.0 \\
MDMMT~\cite{mdmmt}
&& 22.8 & 49.5 & 61.5 & 53.8 & 6.0 \\
MDMMT-2 (Ours)
&& \B{33.3} & \B{59.8} & \B{70.2} & \B{38.7} & \B{3.0} \\
    \bottomrule
  \end{tabular}
\end{table}

\begin{table}
  \centering
  \caption{Test results on LSMDC dataset. Results are collected from articles and \url{https://paperswithcode.com/sota/video-retrieval-on-lsmdc}}
  \label{tab:models-lsmdc}
  
  \begin{tabular}{|l|*{5}l|}
    \toprule
    \multirow{2}{*}{Model} & \multicolumn{5}{c|}{LSMDC text $\rightarrow$ video} \\
			   & R@1$\uparrow$ & R@5$\uparrow$ & R@10$\uparrow$ & MnR$\downarrow$ & MdR$\downarrow$ \\
    \midrule
CT-SAN~\cite{yu2017endtoend}
& 5.1 & 16.3  & 25.2 & --- & 46.0 \\
JSFusion~\cite{yu2018joint}
& 9.1 & 21.2  & 34.1 & --- & 36.0 \\
MEE~\cite{miech2020learning}
& 9.3 & 25.1  & 33.4 & --- & 27.0 \\
MEE-COCO~\cite{miech2020learning}
& 10.1 & 25.6 & 34.6 & --- & 27.0 \\
CE~\cite{liu2020use}
& 11.2 & 26.9 & 34.8 & 96.8 & 25.3 \\
CLIP agg~\cite{portilloquintero2021straightforward}
& 11.3 & 22.7 & 29.2 & --- & 56.5 \\
CLIP~\cite{radford2learning}
& 12.4 & 23.7 & 31.0  & 142.5 & 45.0  \\
MMT ~\cite{gabeur2020multimodal}
& 12.9 & 29.9 & 40.1 & 75.0 & 19.3 \\
MDMMT~\cite{mdmmt}
& 18.8 & 38.5 & 47.9 & 58.0 & 12.3 \\
CLIP4Clip~\cite{CLIP4Clip}
& 21.6 & 41.8 & 49.8 & 58.0 & --- \\
QB-Norm+CLIP4Clip~\cite{qbnorm}
& 22.4 & 40.1 & 49.5 & --- & 11.0 \\
CAMoE~\cite{camoe}
& 25.9 & 46.1 & 53.7 & 54.4 & --- \\
MDMMT-2 (Ours)
& \B{26.9}$_{\pm 0.6}$ & \B{46.7}$_{\pm 0.5}$ & \B{55.9}$_{\pm 0.4}$ & \B{48.0}$_{\pm 0.5}$ & \B{6.7}$_{\pm 0.5}$ \\
    \bottomrule
  \end{tabular}
\end{table}

\begin{table}
  \centering
  \caption{Test results on MSVD dataset. Results are collected from articles and \url{https://paperswithcode.com/sota/video-retrieval-on-msvd}}
  \label{tab:models-msvd}
  
  \begin{tabular}{|l|*{5}l|}
    \toprule
    \multirow{2}{*}{Model} & \multicolumn{5}{c|}{MSVD text $\rightarrow$ video} \\
			   & R@1$\uparrow$ & R@5$\uparrow$ & R@10$\uparrow$ & MnR$\downarrow$ & MdR$\downarrow$ \\
    \midrule
LAFF~\cite{laff}
& 45.4 & 76.0 & 84.6 & --- & --- \\
CLIP4Clip~\cite{CLIP4Clip}
& 46.2 & 76.1 & 84.6 & 10.0 & 2.0 \\
CLIP2Video~\cite{clip2video}
& 47.0 & 76.8 & 85.9 &  9.6 & 2.0 \\
QB-Norm+CLIP2Video~\cite{qbnorm}
& 48.0 & 77.9 & 86.2 & --- & 2.0 \\
CAMoE~\cite{camoe}
& 49.8 & 79.2 & 87.0 & 9.4 & --- \\
MDMMT-2 (Ours)
& \B{56.8}$_{\pm 0.2}$ & \B{83.1}$_{\pm 0.2}$ & \B{89.2}$_{\pm 0.1}$ & \B{8.8}$_{\pm 0.0}$ & \B{1.0}$_{\pm 0.0}$ \\
    \bottomrule
  \end{tabular}
\end{table}

\begin{table}
  \centering
  \caption{Test results on YouCook2 dataset. Results are collected from articles and \url{https://paperswithcode.com/sota/video-retrieval-on-youcook2}}
  \label{tab:models-youcook2}
  
  \begin{tabular}{|l|*{5}l|}
    \toprule
    \multirow{2}{*}{Model} & \multicolumn{5}{c|}{YouCook2  text $\rightarrow$ video} \\
			   & R@1$\uparrow$ & R@5$\uparrow$ & R@10$\uparrow$ & MnR$\downarrow$ & MdR$\downarrow$ \\
    \midrule
Text-Video Embedding~\cite{miech19howto100m}
& 8.2 & 24.5 & 35.3 & --- & 24.0 \\
COOT~\cite{coot}
& 16.7 & --- & 52.3 & --- & --- \\
UniVL~\cite{univl}
& 28.9 & 57.6 & 70.0 &  --- & 4.0 \\
TACo~\cite{yang2021taco}
& 29.6 & 59.7 & 72.7 & --- & 4.0 \\
MDMMT-2 (Ours)
& \B{32.0}$_{\pm 0.7}$ & \B{64.0}$_{\pm 0.3}$ & \B{74.8}$_{\pm 0.2}$ & \B{12.7}$_{\pm 0.3}$ & \B{3.0}$_{\pm 0.0}$ \\
    \bottomrule
  \end{tabular}
\end{table}

\begin{table}
  \centering
  \caption{Test results on TGIF dataset. Results are collected from articles and \url{https://paperswithcode.com/sota/video-retrieval-on-tgif}}
  \label{tab:models-tgif}
  
  \begin{tabular}{|l|*{5}l|}
    \toprule
    \multirow{2}{*}{Model} & \multicolumn{5}{c|}{TGIF  text $\rightarrow$ video} \\
			   & R@1$\uparrow$ & R@5$\uparrow$ & R@10$\uparrow$ & MnR$\downarrow$ & MdR$\downarrow$ \\
    \midrule
W2VV++~\cite{w2vv++}
& 9.4 & 22.3 & 29.8 &  --- & --- \\
SEA~\cite{sea}
& 11.1 & 25.2 & 32.8 & --- & --- \\
LAFF~\cite{laff}
& 24.5 & 45.0 & 54.5 & --- & --- \\
MDMMT-2 (Ours)
& \B{25.5}$_{\pm 0.1}$ & \B{46.1}$_{\pm 0.0}$ & \B{55.7}$_{\pm 0.1}$ & \B{94.1}$_{\pm 0.3}$ & \B{7.0}$_{\pm 0.0}$ \\
    \bottomrule
  \end{tabular}
\end{table}

%% file: sect_conclusions.tex
We performed a refined study of each conceptual part of transformer application for the text-to-video retrieval task. The analysis of the prior knowledge allows to choose optimal existing backbone experts. Combining different types of data sources allows to significantly increase the overall training data amount. Also we suggest a multi-stage training procedure without experts fine-tuning, which prevents their overfitting to a particular domain. Usage of the expanded data and optimal experts leads to a great increase in the generalization ability. It allows to obtain a model, which simultaneously performs well in multiple domains and benefits with the domains diversity increasing. We demonstrate an incredible novelty -- possibility to obtain SOTA results in different domains by a same model, instead of preparing a domain-specific model for each. In particular, we obtained new SOTA results in MSR-VTT, LSMDC, MSVD, YouCook2 and TGIF with a single model trained only once.

%% file: references.bib
@misc{gabeur2020multimodal,
	title={Multi-modal Transformer for Video Retrieval}, 
	author={Valentin Gabeur and Chen Sun and Karteek Alahari and Cordelia Schmid},
	year={2020},
	eprint={2007.10639},
	archivePrefix={arXiv},
	primaryClass={cs.CV}
}

@misc{ghadiyaram2019largescale,
      title={Large-scale weakly-supervised pre-training for video action recognition}, 
      author={Deepti Ghadiyaram and Matt Feiszli and Du Tran and Xueting Yan and Heng Wang and Dhruv Mahajan},
      year={2019},
      eprint={1905.00561},
      archivePrefix={arXiv},
      primaryClass={cs.CV}
}

@misc{yu2018joint,
      title={A Joint Sequence Fusion Model for Video Question Answering and Retrieval}, 
      author={Youngjae Yu and Jongseok Kim and Gunhee Kim},
      year={2018},
      eprint={1808.02559},
      archivePrefix={arXiv},
      primaryClass={cs.CV}
}

@misc{miech2020learning,
      title={Learning a Text-Video Embedding from Incomplete and Heterogeneous Data}, 
      author={Antoine Miech and Ivan Laptev and Josef Sivic},
      year={2020},
      eprint={1804.02516},
      archivePrefix={arXiv},
      primaryClass={cs.CV}
}

@InProceedings{xu2016msr-vtt,
author = {Xu, Jun and Mei, Tao and Yao, Ting and Rui, Yong},
title = {MSR-VTT: A Large Video Description Dataset for Bridging Video and Language},
year = {2016},
publisher = {IEEE International Conference on Computer Vision and Pattern Recognition (CVPR)},
}

@inproceedings{2020trecvidawad,
author= {George Awad and Asad A. Butt and Keith Curtis and Yooyoung Lee and Jonathan Fiscus 
	 and Afzal Godil and Andrew Delgado and Jesse Zhang and Eliot Godard and Lukas 
	 Diduch and Jeffrey Liu and Alan F. Smeaton and Yvette Graham and Gareth J. F. Jones 
	 and Wessel Kraaij and Georges Quénot},
title = {TRECVID 2020:  comprehensive campaign for evaluating video retrieval tasks 
	 across multiple application domains},
booktitle = {Proceedings of TRECVID 2020},
year = 2020,
organization = {NIST, USA},
}

@inproceedings{miech19howto100m,
   title={How{T}o100{M}: {L}earning a {T}ext-{V}ideo {E}mbedding by {W}atching {H}undred {M}illion {N}arrated {V}ideo {C}lips},
   author={Miech, Antoine and Zhukov, Dimitri and Alayrac, Jean-Baptiste and Tapaswi, Makarand and Laptev, Ivan and Sivic, Josef},
   booktitle={ICCV},
   year={2019},
}

@InProceedings{tgif-cvpr2016,
  author = {Li, Yuncheng and Song, Yale and Cao, Liangliang and Tetreault, Joel and Goldberg, Larry and Jaimes, Alejandro and Luo, Jiebo},
  title = "{TGIF: A New Dataset and Benchmark on Animated GIF Description}",
  booktitle = {The IEEE Conference on Computer Vision and Pattern Recognition (CVPR)},
  year = {2016}
}

@misc{rohrbach2016movie,
      title={Movie Description}, 
      author={Anna Rohrbach and Atousa Torabi and Marcus Rohrbach and Niket Tandon and Christopher Pal and Hugo Larochelle and Aaron Courville and Bernt Schiele},
      year={2016},
      eprint={1605.03705},
      archivePrefix={arXiv},
      primaryClass={cs.CV}
}

@inproceedings{chen-dolan-2011-collecting,
    title = "Collecting Highly Parallel Data for Paraphrase Evaluation",
    author = "Chen, David  and
      Dolan, William",
    booktitle = "Proceedings of the 49th Annual Meeting of the Association for Computational Linguistics: Human Language Technologies",
    year = "2011",
    address = "Portland, Oregon, USA",
    publisher = "Association for Computational Linguistics",
    pages = "190--200",
}

@misc{goyal2017something,
      title={The "something something" video database for learning and evaluating visual common sense}, 
      author={Raghav Goyal and Samira Ebrahimi Kahou and Vincent Michalski and Joanna Materzyńska and Susanne Westphal and Heuna Kim and Valentin Haenel and Ingo Fruend and Peter Yianilos and Moritz Mueller-Freitag and Florian Hoppe and Christian Thurau and Ingo Bax and Roland Memisevic},
      year={2017},
      eprint={1706.04261},
      archivePrefix={arXiv},
      primaryClass={cs.CV}
}

@article{radford2learning,
  title={Learning Transferable Visual Models From Natural Language Supervision},
  author={Radford, Alec and Kim, Jong Wook and Hallacy, Chris and Ramesh, Aditya and Goh, Gabriel and Agarwal, Sandhini and Sastry, Girish and Askell, Amanda and Mishkin, Pamela and Clark, Jack and others},
  journal={Image},
  volume={2},
  pages={T2}
}

@inproceedings{imagenet_cvpr09,
        AUTHOR = {Deng, J. and Dong, W. and Socher, R. and Li, L.-J. and Li, K. and Fei-Fei, L.},
        TITLE = {{ImageNet: A Large-Scale Hierarchical Image Database}},
        BOOKTITLE = {CVPR09},
        YEAR = {2009}
}

@inproceedings{mithun2018learning,
  title={Learning joint embedding with multimodal cues for cross-modal video-text retrieval},
  author={Mithun, Niluthpol Chowdhury and Li, Juncheng and Metze, Florian and Roy-Chowdhury, Amit K},
  booktitle={Proceedings of the 2018 ACM on International Conference on Multimedia Retrieval},
  pages={19--27},
  year={2018}
}

@inproceedings{Karpathy2014DeepFE,
author = {Karpathy, Andrej and Joulin, Armand and Fei-Fei, Li},
title = {Deep Fragment Embeddings for Bidirectional Image Sentence Mapping},
year = {2014},
publisher = {MIT Press},
address = {Cambridge, MA, USA},
booktitle = {Proceedings of the 27th International Conference on Neural Information Processing Systems - Volume 2},
pages = {1889–1897},
numpages = {9},
series = {NIPS'14}
}

@misc{patrick2021supportset,
      title={Support-set bottlenecks for video-text representation learning}, 
      author={Mandela Patrick and Po-Yao Huang and Yuki Asano and Florian Metze and Alexander Hauptmann and João Henriques and Andrea Vedaldi},
      year={2021},
      eprint={2010.02824},
      archivePrefix={arXiv},
      primaryClass={cs.CV}
}

@misc{liu2020use,
      title={Use What You Have: Video Retrieval Using Representations From Collaborative Experts}, 
      author={Yang Liu and Samuel Albanie and Arsha Nagrani and Andrew Zisserman},
      year={2020},
      eprint={1907.13487},
      archivePrefix={arXiv},
      primaryClass={cs.CV}
}

@misc{yu2017endtoend,
      title={End-to-end Concept Word Detection for Video Captioning, Retrieval, and Question Answering}, 
      author={Youngjae Yu and Hyungjin Ko and Jongwook Choi and Gunhee Kim},
      year={2017},
      eprint={1610.02947},
      archivePrefix={arXiv},
      primaryClass={cs.CV}
}

@misc{portilloquintero2021straightforward,
      title={A Straightforward Framework For Video Retrieval Using CLIP}, 
      author={Jesús Andrés Portillo-Quintero and José Carlos Ortiz-Bayliss and Hugo Terashima-Marin},
      year={2021},
      eprint={2102.12443},
      archivePrefix={arXiv},
      primaryClass={cs.CV}
}

@article{Dong_2018,
   title={Predicting Visual Features From Text for Image and Video Caption Retrieval},
   volume={20},
   ISSN={1941-0077},
   DOI={10.1109/tmm.2018.2832602},
   number={12},
   journal={IEEE Transactions on Multimedia},
   publisher={Institute of Electrical and Electronics Engineers (IEEE)},
   author={Dong, Jianfeng and Li, Xirong and Snoek, Cees G. M.},
   year={2018},
   pages={3377–3388}
}

@misc{dong2019dual,
      title={Dual Encoding for Zero-Example Video Retrieval}, 
      author={Jianfeng Dong and Xirong Li and Chaoxi Xu and Shouling Ji and Yuan He and Gang Yang and Xun Wang},
      year={2019},
      eprint={1809.06181},
      archivePrefix={arXiv},
      primaryClass={cs.CV}
}

@misc{miech2020endtoend,
      title={End-to-End Learning of Visual Representations from Uncurated Instructional Videos}, 
      author={Antoine Miech and Jean-Baptiste Alayrac and Lucas Smaira and Ivan Laptev and Josef Sivic and Andrew Zisserman},
      year={2020},
      eprint={1912.06430},
      archivePrefix={arXiv},
      primaryClass={cs.CV}
}

@misc{rouditchenko2020avlnet,
      title={AVLnet: Learning Audio-Visual Language Representations from Instructional Videos}, 
      author={Andrew Rouditchenko and Angie Boggust and David Harwath and Dhiraj Joshi and Samuel Thomas and Kartik Audhkhasi and Rogerio Feris and Brian Kingsbury and Michael Picheny and Antonio Torralba and James Glass},
      year={2020},
      eprint={2006.09199},
      archivePrefix={arXiv},
      primaryClass={cs.CV}
}

@misc{kay2017kinetics,
      title={The Kinetics Human Action Video Dataset}, 
      author={Will Kay and Joao Carreira and Karen Simonyan and Brian Zhang and Chloe Hillier and Sudheendra Vijayanarasimhan and Fabio Viola and Tim Green and Trevor Back and Paul Natsev and Mustafa Suleyman and Andrew Zisserman},
      year={2017},
      eprint={1705.06950},
      archivePrefix={arXiv},
      primaryClass={cs.CV}
}

@misc{hershey2017cnn,
      title={CNN Architectures for Large-Scale Audio Classification}, 
      author={Shawn Hershey and Sourish Chaudhuri and Daniel P. W. Ellis and Jort F. Gemmeke and Aren Jansen and R. Channing Moore and Manoj Plakal and Devin Platt and Rif A. Saurous and Bryan Seybold and Malcolm Slaney and Ron J. Weiss and Kevin Wilson},
      year={2017},
      eprint={1609.09430},
      archivePrefix={arXiv},
      primaryClass={cs.SD}
}

@misc{wang2020vatex,
      title={VATEX: A Large-Scale, High-Quality Multilingual Dataset for Video-and-Language Research}, 
      author={Xin Wang and Jiawei Wu and Junkun Chen and Lei Li and Yuan-Fang Wang and William Yang Wang},
      year={2020},
      eprint={1904.03493},
      archivePrefix={arXiv},
      primaryClass={cs.CV}
}

@misc{lei2019tvqa,
      title={TVQA: Localized, Compositional Video Question Answering}, 
      author={Jie Lei and Licheng Yu and Mohit Bansal and Tamara L. Berg},
      year={2019},
      eprint={1809.01696},
      archivePrefix={arXiv},
      primaryClass={cs.CL}
}

@misc{chen2015microsoft,
      title={Microsoft COCO Captions: Data Collection and Evaluation Server}, 
      author={Xinlei Chen and Hao Fang and Tsung-Yi Lin and Ramakrishna Vedantam and Saurabh Gupta and Piotr Dollar and C. Lawrence Zitnick},
      year={2015},
      eprint={1504.00325},
      archivePrefix={arXiv},
      primaryClass={cs.CV}
}

@misc{Sharma2018ConceptualCA,
  title={Conceptual Captions: A Cleaned, Hypernymed, Image Alt-text Dataset For Automatic Image Captioning},
  author={Piyush Sharma and Nan Ding and Sebastian Goodman and Radu Soricut},
  booktitle={ACL},
  year={2018}
}

@misc{young-etal-2014-image,
    title = "From image descriptions to visual denotations: New similarity metrics for semantic inference over event descriptions",
    author = "Young, Peter  and
      Lai, Alice  and
      Hodosh, Micah  and
      Hockenmaier, Julia",
    journal = "Transactions of the Association for Computational Linguistics",
    volume = "2",
    year = "2014",
    address = "Cambridge, MA",
    publisher = "MIT Press",
    doi = "10.1162/tacl_a_00166",
    pages = "67--78",
}

@misc{mdmmt,
   title={MDMMT: Multidomain Multimodal Transformer for Video Retrieval},
   DOI={10.1109/cvprw53098.2021.00374},
   journal={2021 IEEE/CVF Conference on Computer Vision and Pattern Recognition Workshops (CVPRW)},
   publisher={IEEE},
   author={Dzabraev, Maksim and Kalashnikov, Maksim and Komkov, Stepan and Petiushko, Aleksandr},
   year={2021} }

@misc{kazakos2021slowfast,
      title={Slow-Fast Auditory Streams For Audio Recognition}, 
      author={Evangelos Kazakos and Arsha Nagrani and Andrew Zisserman and Dima Damen},
      year={2021},
      eprint={2103.03516},
      archivePrefix={arXiv},
      primaryClass={cs.SD}
}

@misc{CLIP4Clip,
  author  = {Huaishao Luo and Lei Ji and Ming Zhong and Yang Chen and Wen Lei and Nan Duan and Tianrui Li},
  title   = {{CLIP4Clip}: An Empirical Study of CLIP for End to End Video Clip Retrieval},
  journal = {arXiv preprint arXiv:2104.08860},
  year    = {2021},
}

@misc{clip2video,
  title={CLIP2Video: Mastering Video-Text Retrieval via Image CLIP},
  author={Fang, Han and Xiong, Pengfei and Xu, Luhui and Chen, Yu},
  journal={arXiv preprint arXiv:2106.11097},
  year={2021}
}

@misc{camoe,
      title={Improving Video-Text Retrieval by Multi-Stream Corpus Alignment and Dual Softmax Loss}, 
      author={Xing Cheng and Hezheng Lin and Xiangyu Wu and Fan Yang and Dong Shen},
      year={2021},
      eprint={2109.04290},
      archivePrefix={arXiv},
      primaryClass={cs.CV}
}

@misc{gao2021clip2tv,
      title={CLIP2TV: An Empirical Study on Transformer-based Methods for Video-Text Retrieval}, 
      author={Zijian Gao and Jingyu Liu and Sheng Chen and Dedan Chang and Hao Zhang and Jinwei Yuan},
      year={2021},
      eprint={2111.05610},
      archivePrefix={arXiv},
      primaryClass={cs.CV}
}

@misc{qbnorm,
      title={Cross Modal Retrieval with Querybank Normalisation}, 
      author={Simion-Vlad Bogolin and Ioana Croitoru and Hailin Jin and Yang Liu and Samuel Albanie},
      year={2021},
      eprint={2112.12777},
      archivePrefix={arXiv},
      primaryClass={cs.CV}
}

@misc{laff,
      title={Lightweight Attentional Feature Fusion for Video Retrieval by Text}, 
      author={Fan Hu and Aozhu Chen and Ziyue Wang and Fangming Zhou and Xirong Li},
      year={2021},
      eprint={2112.01832},
      archivePrefix={arXiv},
      primaryClass={cs.MM}
}

@misc{yang2021taco,
      title={TACo: Token-aware Cascade Contrastive Learning for Video-Text Alignment}, 
      author={Jianwei Yang and Yonatan Bisk and Jianfeng Gao},
      year={2021},
      eprint={2108.09980},
      archivePrefix={arXiv},
      primaryClass={cs.CV}
}

@misc{gpt,
      title={Language Models are Few-Shot Learners}, 
      author={Tom B. Brown and Benjamin Mann and Nick Ryder and Melanie Subbiah and Jared Kaplan and Prafulla Dhariwal and Arvind Neelakantan and Pranav Shyam and Girish Sastry and Amanda Askell and Sandhini Agarwal and Ariel Herbert-Voss and Gretchen Krueger and Tom Henighan and Rewon Child and Aditya Ramesh and Daniel M. Ziegler and Jeffrey Wu and Clemens Winter and Christopher Hesse and Mark Chen and Eric Sigler and Mateusz Litwin and Scott Gray and Benjamin Chess and Jack Clark and Christopher Berner and Sam McCandlish and Alec Radford and Ilya Sutskever and Dario Amodei},
      year={2020},
      eprint={2005.14165},
      archivePrefix={arXiv},
      primaryClass={cs.CL}
}

@misc{bert,
  author    = {Jacob Devlin and
               Ming{-}Wei Chang and
               Kenton Lee and
               Kristina Toutanova},
  title     = {{BERT:} Pre-training of Deep Bidirectional Transformers for Language
               Understanding},
  journal   = {CoRR},
  volume    = {abs/1810.04805},
  year      = {2018},
  eprinttype = {arXiv},
  eprint    = {1810.04805},
  bibsource = {dblp computer science bibliography, https://dblp.org}
}

@misc{caba2015activitynet,
  title={ActivityNet: A Large-Scale Video Benchmark for Human Activity Understanding},
  author={Fabian Caba Heilbron, Victor Escorcia, Bernard Ghanem and Juan Carlos Niebles},
  booktitle={Proceedings of the IEEE Conference on Computer Vision and Pattern Recognition},
  pages={961--970},
  year={2015}
}

@inproceedings{ZhXuCoAAAI18,
    author={Zhou, Luowei and Xu, Chenliang and Corso, Jason J},
    title = {Towards Automatic Learning of Procedures From Web Instructional Videos},
    booktitle = {AAAI Conference on Artificial Intelligence},
    pages={7590--7598},
    year = {2018},
}

@misc{radford2021learning,
      title={Learning Transferable Visual Models From Natural Language Supervision}, 
      author={Alec Radford and Jong Wook Kim and Chris Hallacy and Aditya Ramesh and Gabriel Goh and Sandhini Agarwal and Girish Sastry and Amanda Askell and Pamela Mishkin and Jack Clark and Gretchen Krueger and Ilya Sutskever},
      year={2021},
      eprint={2103.00020},
      archivePrefix={arXiv},
      primaryClass={cs.CV}
}

@inproceedings{NIPS2016_6b180037,
 author = {Sohn, Kihyuk},
 booktitle = {Advances in Neural Information Processing Systems},
 editor = {D. Lee and M. Sugiyama and U. Luxburg and I. Guyon and R. Garnett},
 pages = {},
 publisher = {Curran Associates, Inc.},
 title = {Improved Deep Metric Learning with Multi-class N-pair Loss Objective},
 volume = {29},
 year = {2016}
}

@article{oord,
  author    = {A{\"{a}}ron van den Oord and
               Yazhe Li and
               Oriol Vinyals},
  title     = {Representation Learning with Contrastive Predictive Coding},
  journal   = {CoRR},
  volume    = {abs/1807.03748},
  year      = {2018},
  eprinttype = {arXiv},
  eprint    = {1807.03748},
  bibsource = {dblp computer science bibliography, https://dblp.org}
}

@article{infoNCE,
  author    = {Richard Zhang},
  title     = {Making Convolutional Networks Shift-Invariant Again},
  journal   = {CoRR},
  volume    = {abs/1904.11486},
  year      = {2019},
  eprinttype = {arXiv},
  eprint    = {1904.11486},
  bibsource = {dblp computer science bibliography, https://dblp.org}
}

@article{miech2018learning,
  title={Learning a text-video embedding from incomplete and heterogeneous data},
  author={Miech, Antoine and Laptev, Ivan and Sivic, Josef},
  journal={arXiv preprint arXiv:1804.02516},
  year={2018}
}

@article{univl,
  author    = {Huaishao Luo and
               Lei Ji and
               Botian Shi and
               Haoyang Huang and
               Nan Duan and
               Tianrui Li and
               Xilin Chen and
               Ming Zhou},
  title     = {UniViLM: {A} Unified Video and Language Pre-Training Model for Multimodal
               Understanding and Generation},
  journal   = {CoRR},
  volume    = {abs/2002.06353},
  year      = {2020},
  eprinttype = {arXiv},
  eprint    = {2002.06353},
  bibsource = {dblp computer science bibliography, https://dblp.org}
}

@article{coot,
  author    = {Simon Ging and
               Mohammadreza Zolfaghari and
               Hamed Pirsiavash and
               Thomas Brox},
  title     = {{COOT:} Cooperative Hierarchical Transformer for Video-Text Representation
               Learning},
  journal   = {CoRR},
  volume    = {abs/2011.00597},
  year      = {2020},
  eprinttype = {arXiv},
  eprint    = {2011.00597},
  bibsource = {dblp computer science bibliography, https://dblp.org}
}

@misc{w2vv++,
author = {Li, Xirong and Xu, Chaoxi and Yang, Gang and Chen, Zhineng and Dong, Jianfeng},
year = {2019},
title = {W2VV++: Fully Deep Learning for Ad-hoc Video Search},
isbn = {978-1-4503-6889-6},
doi = {10.1145/3343031.3350906}
}

@article{sea,
  author={Li, Xirong and Zhou, Fangming and Xu, Chaoxi and Ji, Jiaqi and Yang, Gang},
  journal={IEEE Transactions on Multimedia}, 
  title={SEA: Sentence Encoder Assembly for Video Retrieval by Textual Queries}, 
  year={2021},
  volume={23},
  number={},
  pages={4351-4362},
  doi={10.1109/TMM.2020.3042067}}
